\def\year{2022}\relax
\documentclass[letterpaper]{article}
\usepackage{aaai22} 
\usepackage{times} 
\usepackage{helvet} 
\usepackage{courier} 
\usepackage[hyphens]{url} 
\usepackage{graphicx} 
\usepackage{graphicx} 
\usepackage{natbib} 
\usepackage{filecontents}
\usepackage{caption} 
\DeclareCaptionStyle{ruled}%
{labelfont=normalfont,labelsep=colon,strut=off}
\usepackage{booktabs}       
\usepackage{amsfonts}       
\usepackage{nicefrac}       
\usepackage{microtype}      
\usepackage{amsmath}
\usepackage{amssymb}
\usepackage{amsthm}
\usepackage{bm}
\usepackage{algorithm}
\usepackage{subcaption}
\usepackage{systeme}
\usepackage{bbm}
\newcommand*\diff{\mathop{}\!\mathrm{d}}
\newtheorem{theorem}{Theorem}

\title{Block-wise Training of Residual Networks via the Minimizing Movement Scheme}
\date{}
\author{Skander Karkar,\textsuperscript{\rm 1 2} Ibrahim Ayed,\textsuperscript{\rm 1 3} Emmanuel de Bézenac,\textsuperscript{\rm 1} Patrick Gallinari\textsuperscript{\rm 1 2}}
\affiliations{\textsuperscript{\rm 1}LIP6, Sorbonne Université, France \\
\textsuperscript{\rm 2}Criteo AI Lab, Criteo, France \\
\textsuperscript{\rm 3}Theresis lab, Thales, France \\
\{skander.karkar, ibrahim.ayed, emmanuel.de-bezenac, patrick.gallinari\}@lip6.fr
}

\begin{document}

\maketitle

\begin{abstract}
End-to-end backpropagation has a few shortcomings: it requires loading the entire model during training, which can be impossible in constrained settings, and suffers from three locking problems (forward locking, update locking and backward locking), which prohibit training the layers in parallel. Solving layer-wise optimization problems can address these problems and has been used in on-device training of neural networks. We develop a layer-wise training method, particularly well-adapted to ResNets, inspired by the minimizing movement scheme for gradient flows in distribution space. The method amounts to a kinetic energy regularization of each block that makes the blocks optimal transport maps and endows them with regularity. It works by alleviating the stagnation problem observed in layer-wise training, whereby greedily-trained early layers overfit and deeper layers stop increasing test accuracy after a certain depth. We show on classification tasks that the test accuracy of block-wise trained ResNets is improved when using our method, whether the blocks are trained sequentially or in parallel.
\end{abstract}

\section{Introduction}

End-to-end backpropagation is the standard training method of neural nets. But there are reasons to look for alternatives. It is considered biologically unrealistic \cite{local-errors} and it requires loading the whole model during training which can be impossible in constrained settings such as training on mobile devices \cite{sensors1,sensors2}. It also prohibits training layers in parallel as it suffers from three locking problems (forward locking: each layer must wait for the previous layers to process its input, update locking: each layer must wait for the end of the forward pass to be updated, and backward locking: each layer must wait for errors to backpropagate from the last layer to be updated) \cite{dni1}. These locking problems force the training and deployment of networks to be sequential and synchronous, and breaking them would allow for more flexibility when using networks that are distributed between a central agent and clients and that operate at different rates \cite{dni1}. Greedily solving layer-wise optimization problems, sequentially (i.e. one after the other) or in parallel (i.e. batch-wise), solves update locking (and so also backward locking). When combined with buffers, parallel layer-wise training solves all three problems \cite{belilovsky2} and allows distributed training of the layers. Layer-wise training is appealing in memory-constrained settings as it works without most gradients and activations needed in end-to-end training, and when done sequentially, only requires loading and training one layer at a time. Despite its simplicity, layer-wise training has been shown \cite{belilovsky1,belilovsky2} to scale well. It outperforms more complicated ideas developed to address the locking problems such as synthetic \cite{dni1,dni2} and delayed \cite{ddg,features-replay} gradients. We can also deduce theoretical results about a network of greedily-trained shallow sub-modules from the theoretical results about shallow networks \cite{belilovsky1,belilovsky2}. Module-wise training, where the network is split into modules that are trained greedily, may not offer the same computational gains as full layer-wise training, but it gets closer to the accuracy of end-to-end training and is explored often \cite{sedona,infopro}.

The typical setting of (sequential) module-wise training for minimizing a loss $L$, is, given a dataset $\Tilde{\rho}_0$, to solve one after the other, for $0 \leq k \leq K$, the problems
\begin{equation}
    \label{eq:layerwise-problem}
    (T_k,F_k) \in \arg \min_{T,F} \sum_{x \in \Tilde{\rho}_0} L(F, T(G_{k-1}(x))
\end{equation}
where $G_k = T_k \circ ... \circ T_0$ for $0 \leq k \leq K$ and $G_{-1} = \texttt{id}$. Here, $T_k$ is the module (a single layer or a group of layers) and $F_k$ is an auxiliary network (a classifier if the task is classification) that processes the outputs of $T_k$ so that the loss can be computed. In a generation task, auxiliary networks might not be needed. In all cases, module $T_{k+1}$ receives the output of module $T_k$. The final network trained this way is $F_K \circ G_K$, but we can stop at any previous depth $k$ and use $F_k \circ G_k$ if it performs better. In fact, and especially when modules are shallow, module-wise training suffers from a stagnation problem, whereby greedily-trained early modules overfit and deeper modules stop improving the test accuracy after a certain depth, or even degrade it \cite{cascade,infopro}. We observe this experimentally (Figure \ref{fig:seq-par}) and propose a regularization for module-wise training that addresses this problem by increasing training stability. The regularization is particularly well-adapted to ResNets \cite{resnet2,resnet1} and similar models such as ResNeXts \cite{resnext} and Wide ResNets \cite{wide}, but is easily usable on other models, and ResNets themselves remain competitive \cite{touvron}. The method leverages the analogy between ResNets and the Euler scheme for ODEs \cite{weinan} to penalize the kinetic energy of the network. Intuitively, if the kinetic energy is penalized enough, the current module will barely move the points, thus at least preserving the accuracy of the previous module and avoiding its collapse.


After a discussion of related work in Section 2, we present the method in Section 3 and show that it amounts to a transport regularization of each module, which we prove forces the solution module to be an optimal transport map and makes it regular. This also suggests a simple principled extension of the method to non-residual networks. In Section 3, we link the method with gradient flows and the minimizing movement scheme in the Wasserstein space, which allows to invoke convergence results to a minimizer under additional hypotheses on the loss. Section 4 discusses different practical implementations and introduces a new variant of layer-wise training we call \textit{multi-lap sequential} training. It is a slight variation on sequential layer-wise training that has the same advantages and offers a non-negligible improvement in many cases over sequential training for the same computational and memory costs. Experiments using different architectures and on different classification datasets in Section 5 show that our method consistently improves the test accuracy of block-wise and module-wise trained residual networks, particularly in small data regimes, whether the block-wise training is carried out sequentially or in parallel.

\section{Related work}

Layer-wise training of neural networks has been considered as a pre-training and initialization method \cite{bengio,cascade} and was shown recently to perform competitively with end-to-end training \cite{belilovsky1,nokland}. This has led to it being considered in practical settings with limited resources such as embedded training \cite{sensors1,sensors2}. For layer-wise training, many papers consider using a different auxiliary loss, instead of or in addition to the classification loss: kernel similarity \cite{kernel-similarity}, information-theory-inspired losses \cite{gim,information-bottleneck,hsic,infopro} and biologically plausible losses \cite{gim,nokland,bio,clapp,loco}. Paper \cite{belilovsky1} reports the best experimental results when solving the layer-wise problems sequentially. Methods PredSim \cite{nokland}, DGL \cite{belilovsky2}, Sedona \cite{sedona} and InfoPro \cite{infopro} report the best results when solving the layer-wise problems in parallel, albeit each in a somewhat different setting. \cite{belilovsky1,belilovsky2} do it simply through architectural considerations mostly regarding the auxiliary networks. However, \cite{belilovsky1} do not consider ResNets and PredSim state that their method does not perform well on ResNets, specifically because of the skip connections. DGL only considers a ResNet architecture by splitting it in the middle and training the two halves without backpropagating between them. All three focus on VGG architectures and networks that are not deep. Sedona applies architecture search to decide on where to split the network into 2 or 4 modules and what auxiliary classifier to use before module-wise training. Only BoostResNet \cite{boosting} also proposes a block-wise training idea geared for ResNets. However, their results only show better early performance on limited experiments and end-to-end fine-tuning is required to be competitive. A method called ResIST \cite{resist} that is similar to block-wise training of ResNets randomly assigns residual blocks to one of up to 8 sub-networks that are trained independently and reassembled before another random partition. But only blocks in the third section of the ResNet are partitioned, the same block can appear in many sub-networks and the sub-networks are not necessarily made up of successive blocks. Considered as a distributed training method, it is only compared with local SGD \cite{local-sgd}. These methods can all be combined with our regularization, and we use the auxiliary network architecture from \cite{belilovsky1,belilovsky2}. We also show the benefits of our method both with full layer-wise training and when the network is split into a few modules.

Besides layer-wise training, methods such as DNI \cite{dni1,dni2}, DDG \cite{ddg} and Features Replay \cite{features-replay}, solve the update locking problem and the backward locking problem with an eye towards parallelization by using delayed or synthetic predicted gradients, or even predicted inputs to address forward locking. But they only fully apply this to quite shallow networks and only split deeper ones into a small number of sub-modules (less than five) that don't backpropagate to each other and observe training issues with more splits \cite{features-replay}. This makes them compare unfavorably to layer-wise training \cite{belilovsky2}. The high dimension of the predicted gradient which scales with the size of the network renders \cite{dni1,dni2} challenging in practice. Therefore, despite its simplicity, greedy layer-wise training is more appealing when working in a constrained setting. 

Viewing residual networks as dynamical transport systems \cite{oudt,lap} followed from their view as a discretization of differential equations \cite{weinan,lm}. Transport regularization was also used in \cite{nodetrain} to accelerate the training of the NeuralODE model \cite{node}. Transport regularization of ResNets in particular is motivated by the observation that they are naturally biased towards minimally modifying  their input \cite{iterative,hauser,lap}. We further link this transport viewpoint with gradient flows in the Wasserstein space to apply it in a principled way to module-wise training. Gradient flows in the Wasserstein space operating on the data space appeared recently in deep learning. In \cite{alvarez2}, the focus is on functionals of measures whose first variations are known in closed form and used, through their gradients, in the algorithm. This limits the scope of their applications to transfer learning and similar tasks. Likewise, \cite{vgrow,slicedflow,mmdflow,dgflow}
use the explicit gradient flow of $f$-divergences and other distances between measures for generation and generator refinement. In contrast, we use the minimizing movement scheme which does not require computation of the first variation and allows to consider classification. 

\section{Regularized block-wise training of ResNets}

In this section we state the module-wise problems we solve and show that the modules they induce are regular as they approximate optimal transport maps. We show that solving these problems sequentially means following a minimizing movement that approximates the Wasserstein gradient flow that minimizes the loss, which offers hints as to why it works well in practice outside the theoretical hypotheses.

\subsection{Method statement}

In a ResNet, a data point $x_0$ is transported by applying $x_{m+1} = x_m + g_m(x_m)$ for $M$ ResBlocks (a ResBlock is a function $\texttt{id} + g_m$), and $x_{M+1}$ is then classified. To keep the greedily-trained modules from overfitting and destroying information needed by deeper modules, we propose to penalize their kinetic energy to force them to preserve the geometry of the problem as much as possible. The total discrete kinetic energy of a ResNet (for a single point $x_0$) is $\sum \|g_m(x_m)\|^2$, since a ResNet can be seen as an Euler scheme for an ODE  with velocity field $g$ \cite{weinan}:
\begin{equation}
    \label{eq:resnet-euler}
    x_{m+1} = x_m + g_m(x_m) \; \longleftrightarrow \; \partial_t x_t = g_t(x_t)
\end{equation}
That ResNets are already biased towards small displacements and therefore low kinetic energy and that this bias is desirable and should be encouraged has been observed in many works \cite{iterative,small-step,hauser,batchnorm,lap}. Using the notations from \eqref{eq:layerwise-problem}, if each module $T_k$ is made up of $M$ ResBlocks, i.e. has the form $(\texttt{id} + g_{M-1}) \circ ... \circ (\texttt{id} + g_0)$, we propose to penalize its kinetic energy over its input points by adding it to the loss $L$ in the target of the greedy problems \eqref{eq:layerwise-problem}. We denote $\psi_m^x$ the position of an input $x$ after $m$ ResBlocks. Given $\tau > 0$ used to weight the regularization, we solve, for $0 \leq k \leq K$, problems 
\begin{align}
    \label{eq:regularized-layerwise-problem}
    (T_k^\tau, F_k^\tau) & \in \\ \nonumber
    \arg\min_{T,F} & \sum_{x \in \Tilde{\rho}_0} ( L(F, T(G_{k-1}^\tau(x)) + \frac{1}{2 \tau} \sum_{m=0}^{M-1} \| g_m(\psi_m^x) \|^2 ) \\ \nonumber
    & \text{s.t. } T = (\texttt{id} + g_{M-1}) \circ ... \circ (\texttt{id} + g_0) \\ \nonumber
    & \ \ \ \ \psi_0^x = G_{k-1}^\tau(x), \psi_{m+1}^x = \psi_m^x + g_m(\psi_m^x)\  \nonumber
\end{align}
where $G^\tau_k = T^\tau_k \circ ... \circ T^\tau_0$ for $0 \leq k \leq K$ and $G^\tau_{-1} = \texttt{id}$. The final network is now $F^\tau_K \circ G^\tau_K$. Intuitively, we can think that this biases the modules towards moving the points as little as possible, thus at least keeping the performance of the previous module. In our experiments, we will mostly focus on block-wise training, i.e. the case $M=1$ where each $T^\tau_k$ is a single residual block, as it is more challenging.  

\subsection{Regularity result}\label{subsec:regularity}

The Appendix gives the necessary background on optimal transport (OT) theory to prove a regularity result for our method. We start by moving to a continuous viewpoint. We denote $\rho_0 = \rho^\tau_0$ the data distribution of which $\Tilde{\rho}_0$ is a sample and $\mathcal{L}$ the distribution-wide loss that arises from the point-wise loss $L$. As expressed in \eqref{eq:resnet-euler}, a residual network can be seen as an Euler discretization of a differential equation. Problem \eqref{eq:regularized-layerwise-problem} is then the discretization of problem 
\begin{align} 
    \label{eq:nn-wass-min-movement-joint-dyn}
    (T_k^{\tau}, F_k^{\tau}) & \in \\ \nonumber 
    \arg\min_{T,F} & \ \mathcal{L}(F, T_\sharp \rho_k^{\tau}) + \frac{1}{2 \tau} \int_0^1 \|v_t\|_{L^2((\phi^\cdot_t)_\sharp \rho_k^{\tau})}^2 \diff t \\ \nonumber
    & \text{s.t. } T = \phi_1^\cdot, \ \partial_t\phi_t^x = v_t(\phi_t^x), \ \phi^\cdot_0 = \text{id} \nonumber
\end{align}
where $\rho_{k+1}^{\tau} = (T_k^{\tau})_\sharp \rho_k^{\tau}$ and $g_m$ is the discretization of vector field $v_t$ at time $t=m/M$. Here, data distributions $\rho_k^{\tau}$ are pushed forward through the maps $T_k^{\tau}$ which correspond to the flow at $t=1$ of the kinetically-regularized velocity field $v_t$. Given the equivalence between the Monge OT problem \eqref{eq:monge} and the OT problem in dynamic form \eqref{eq:brenier2} in the Appendix, problem \eqref{eq:nn-wass-min-movement-joint-dyn} is equivalent to
\begin{align}
    \label{eq:nn-wass-min-movement-joint}
    (T_k^{\tau}, F_k^{\tau}) & \in \\ \nonumber 
    \arg\min_{T,F} & \ \mathcal{L}(F, T_\sharp \rho_k^{\tau}) + \frac{1}{2 \tau} \int_{\Omega} \| T(x) - x \|^2 \diff \rho_k^{\tau}(x) \nonumber 
\end{align} 
where points are moved instantly through $T$ instead of infinitesimally through velocity filed $v$. This equivalent formulation leads to another discretization \eqref{eq:nn-wass-min-movement-discrete} and implementation of the method more easily applicable in practice to non-residual architectures by simply penalizing the difference between the module's output and its input. We can show that problem \eqref{eq:nn-wass-min-movement-joint} indeed has a solution and that $T_k^{\tau}$ is necessarily an optimal transport between its input and output distributions, which means that it comes with some regularity. We assume that the minimization in $F$ is over a compact set $\mathcal{F}$, that $\rho_k^\tau$ is absolutely continuous, that $\mathcal{L}$ is continuous and non-negative and that $\Omega$ is compact.

\begin{theorem}
\label{th:existence} Problems \eqref{eq:nn-wass-min-movement-joint} and \eqref{eq:nn-wass-min-movement-joint-dyn} have a minimizer $(T_k^{\tau}, F_k^{\tau})$ such that $T_k^{\tau}$ is an optimal transport map. And for any minimizer $(T_k^{\tau}, F_k^{\tau})$, $T_k^{\tau}$ is an optimal transport map.
\end{theorem}
The proof is in the Appendix. Optimal transport maps have regularity properties under some boundedness assumptions. Given Theorem \ref{th:regularity} in the Appendix taken from \cite{figalli}, $T_k^{\tau}$ is $\eta$-Hölder continuous almost everywhere and if the optimization algorithm we use to solve the discretized problem \eqref{eq:regularized-layerwise-problem} returns an approximate solution pair $(\Tilde{F}_k^{\tau}, \Tilde{T}_k^{\tau})$ such that $\Tilde{T}_k^{\tau}$ is an $\epsilon$-optimal transport map, i.e. $\| \Tilde{T}_k^{\tau} - T_k^{\tau} \|_\infty \leq \epsilon$, then we have (using the triangle inequality) the following stability property of the neural module $\Tilde{T}_k^{\tau}$:
\begin{equation}
\label{eq:stability}
\| \Tilde{T}_k^{\tau}(x) - \Tilde{T}_k^{\tau}(y) \| \leq 2 \epsilon + C \| x - y \|^\eta
\end{equation}
for almost every  $x,y \in \text{supp}(\rho_k^\tau)$ and a constant $C>0$. The experimental advantages of using such networks have been shown in \cite{lap}. Naively composing these stability bounds on $T_k^{\tau}$ and $\Tilde{T}_k^{\tau}$ allows to get stability bounds for the composition networks $G^\tau_K$ and $\Tilde{G}^\tau_K = \Tilde{T}^\tau_K \circ .. \circ \Tilde{T}^\tau_0$.

\subsection{Link with the minimizing movement scheme}

The Appendix gives a background on gradient flows and the minimizing movement scheme following \cite{ambrosio,santambrogio2016}. Given a compact set $\Omega \subset \mathbb{R}^d$ and a lower semi-continuous function $\mathcal{L}: \mathbb{W}_2(\Omega) \rightarrow \mathbb{R} \cup \{ \infty \}$, the minimizing movement scheme is a discretized gradient flow that is well-defined in non-Euclidean metric spaces and can, under some conditions, minimize $\mathcal{L}$ starting from $\rho^\tau_0 \in \mathcal{P}(\Omega)$. It is given by
\begin{equation}
\label{eq:wass-min-movement1}
    \rho^{\tau}_{k+1} \in \arg \min_{\rho \in \mathcal{P}(\Omega)} \ \mathcal{L}(\rho) + \frac{1}{2 \tau} W_2^2(\rho, \rho^{\tau}_k)
\end{equation}
This problem has a solution because the objective is lower semi-continuous and $\mathcal{P}(\Omega)$ is compact. It is equivalent to 
\begin{equation}
     T^\tau_{k+1} \in \arg \min_{T:\Omega \rightarrow \Omega} \mathcal{L}(T_\sharp \rho^{\tau}_k) + \frac{1}{2 \tau} W_2^2(T_\sharp \rho^{\tau}_k, \rho^{\tau}_k)
    \label{eq:wass-min-movement2}
\end{equation}
with $\rho_{k+1}^{\tau} = (T_k^{\tau})_\sharp \rho_k^{\tau}$, under conditions that guarantee the existence of a transport map between $\rho_k^{\tau}$ and any other measure, for example $\partial \Omega$ negligible and $\rho_k^{\tau}$ absolutely continuous, and $\mathcal{L}$ can ensure that $\rho_{k+1}^{\tau}$ is also absolutely continuous. Among the functions $T^\tau_{k+1}$ that solve problem \eqref{eq:wass-min-movement2}, is the optimal transport map from $\rho_k^{\tau}$ to $\rho_{k+1}^{\tau}$. To solve for this optimal map, we consider the following equivalent problem
\begin{equation}
    T^\tau_{k+1} \in \arg \min_{T: \Omega \rightarrow \Omega} \ \mathcal{L}(T_\sharp \rho^{\tau}_k) + \frac{1}{2 \tau} \int_{\Omega} \| T(x) - x \|^2 \diff \rho^{\tau}_k(x)
    \label{eq:wass-min-movement3}
\end{equation}
This problem is equivalent to problem \eqref{eq:wass-min-movement2} as it has the same minimum value, but its minimizer is now an optimal transport map between $\rho_k^{\tau}$ and a minimizer $\rho_{k+1}^{\tau}$ of \eqref{eq:wass-min-movement1}. 

Therefore, if we cast a learning problem as the minimization of a loss $\mathcal{L}$ over the space of measures starting from the input data distribution $\rho_0$, then this suggests solving it through the minimizing movement scheme in $\mathbb{W}_2(\Omega)$, using neural nets $T_k^{\tau}$ found by solving \eqref{eq:wass-min-movement3} that will have regularity \eqref{eq:stability} to push the successive distributions $\rho^{\tau}_k$ towards a minimizer of $\mathcal{L}$. We find again our regularization for module-wise training of neural nets as formulated in \eqref{eq:nn-wass-min-movement-joint}, when auxiliary networks are not necessary, for example in generative tasks. Note that the representation power of neural nets shown by universal approximation theorems is important here to get close to equivalence between \eqref{eq:wass-min-movement1} and \eqref{eq:wass-min-movement2} when restricting the optimization in \eqref{eq:wass-min-movement2} to neural nets.

As mentioned in the Appendix, for losses $\mathcal{L}$ that are $\lambda$-geodesically convex for $\lambda>0$, we can show convergence of this scheme as $k \rightarrow \infty$ and $\tau \rightarrow 0$ to a minimizer of $\mathcal{L}$, potentially under more technical conditions \cite{santambrogio2016}. Functionals of distributions that might be $\lambda$-geodesically convex and are useful as machine learning losses (mostly for generation and transfer tasks) include $\mathcal{V}(\rho) = \int V \diff \rho$ (depending on real-valued function $V$) \cite{santambrogio2016}). Since these functionals have been explored in \cite{alvarez2}, and since their first variations are known in closed form, rendering optimization methods other than the minimizing movement scheme applicable, we choose to focus on less amenable (from this viewpoint) tasks, namely classification. In tasks such as classification, the loss is preceded by a function (a classification head) that maps the transported input to the desired dimension. Introducing this function into the minimizing movement scheme \eqref{eq:wass-min-movement3} leads directly to our regularized module-wise training problem as formulated in \eqref{eq:nn-wass-min-movement-joint}. This is no longer exactly a minimizing movement scheme, except in applications where we can fix the functions $F^\tau_k = F$, but the convergence discussion still suggests taking $\tau$ as small as possible and many modules $T^\tau_k$.

\section{Practical implementation}

The module-wise problems \eqref{eq:layerwise-problem} can be solved in one of two ways. One can completely train each module with its auxiliary classifier for $N$ epochs before training the next module, which receives as input the output of the previous trained module. We call this \textit{sequential} module-wise training. But we can also do this batch-wise, i.e. do a complete forward pass on each batch but without a full backward pass, rather a backward pass that only updates the current module $T_k^\tau$ and its auxiliary classifier $F_k^\tau$, meaning that $T_k^\tau$ forwards its output to $T_{k+1}^\tau$ immediately after it computes it. We call this \textit{parallel} module-wise training. It is called \textit{decoupled} greedy training in \cite{belilovsky2}, which shows that combining it with buffers solves all three locking problems and allows a linear training parallelization in the depth of the network. We propose a variant of sequential module-wise training that we call \textit{multi-lap sequential} module-wise training, in which instead of training each module for $N$ epochs, we train each module from the first to the last sequentially for $N/R$ epochs, then go back and train from the first module to the last for $N/R$ epochs again, and we do this for $R$ laps. For the same total number of epochs and training time, and the same advantages (loading and training one module at a time) this provides a non-negligible improvement in accuracy over normal sequential module-wise training in most cases, as shown below. Despite our theoretical framework being that of sequential module-wise training, our method improves the test accuracy of all three layer-wise training regimes. And in all three cases, we use SGD. 

Formulation \eqref{eq:nn-wass-min-movement-joint} gives another way of implementing the regularization, especially useful for non-residual architecture if the module preserves the dimension. Its discretization is
\begin{align}
    \label{eq:nn-wass-min-movement-discrete}
    (T_k^\tau, F_k^\tau) \in \arg\min_{T,F} & \sum_{x \in \Tilde{\rho}_0} L(F,T(G_{k-1}^\tau(x))  + \\ \nonumber 
    & + \frac{1}{2 \tau} \| T(G_{k-1}^\tau(x)) - G_{k-1}^\tau(x) \|^2 \ \nonumber
\end{align}

Simple variations of the method work better in practice in some cases. For example, instead of using a fixed weight $\tau$ for the transport cost, we can vary it along the depth $k$ to further constrain with a smaller $\tau_k$ the earlier modules to avoid that they overfit or the later modules to maintain the performance of earlier modules.

We might also want to regularize the networks further in earlier epochs when the data is more entangled as in \cite{lap}. To unify and formalize this varying weight $\tau_{k,i}$ across modules $k$ and SGD iterations $i$, we use a scheme inspired by the method of multipliers to solve problems \eqref{eq:regularized-layerwise-problem} and \eqref{eq:nn-wass-min-movement-discrete}. To simplify the notations, we will instead consider the weight $\lambda_{k,i} = 2 \tau_{k,i}$ given to the loss. We denote $\theta_{k,i}$ the parameters of both $T_k$ and $F_k$ at SGD iteration $i$. We also denote $L(\theta,x)$ and $T(\theta,x)$ respectively the loss and the transport regularization as functions of parameters $\theta$ and data point $x$. We now increase the weight $\lambda_{k,i}$ of the loss every $s$ iterations of SGD by a value that is proportional to the current loss. Given increase factor $h > 0$, initial parameters $\theta_{k,1}$, initial weight $\lambda_{k,1} \geq 0$, learning rates $(\eta_i)$ and batches $(x_i)$, we apply for module $k$ and $i \geq 1$:
\begin{align*}
\label{eq:uzawa}
\theta_{k,i+1} &= \theta_{k,i} - \eta_i \nabla_\theta (\lambda_{k,i} \ L(\theta_{k,i},x_i) + T(\theta_{k,i},x_i))  \\
\lambda_{k,i+1} &= \lambda_{k,i} +  h L(\theta_{k,i+1}, x_{i+1}) \text{ if } i \text{ mod } s = 0 \text{ else } \lambda_{k,i}
\end{align*}
The weights $\lambda_{k,i}$ will vary along the depth $k$ even if we use the same initial weights $\lambda_{k,1} = \lambda_1$ because they will evolve differently with iterations $i$ for each $k$. They will increase more slowly with $i$ for larger $k$ because deeper modules will have smaller loss. We are then regularizing the earlier blocks less. This method can be seen as a method of multipliers for the problem of minimizing the transport under the constraint of zero loss (a reasonable assumption as recent deep learning architectures have shown to systematically achieve near zero training loss \cite{generalization,ntk,belkin2018,belkin2019}). Therefore it is immediate by slightly adapting the proof of Theorem \ref{th:existence} or from \cite{lap} that we are still solving a problem that admits a solution whose non-auxiliary part is an optimal transport map with the same regularity as stated above. This method works better than a simple fixed $\tau$ in some experiments, but has more hyperparameters to be tuned.

\section{Experiments}
\renewcommand{\arraystretch}{1.5}

We consider classification tasks, with $L$ being the cross-entropy loss. For the ResBlocks, we use the architecture from \cite{resnet2}. For the auxiliary classifiers, we use the architecture from \cite{belilovsky1,belilovsky2}, that is a convolution followed by an average pooling and a fully connected layer. The experiments below then show that our method combines well with theirs and improves on it when using ResNets. Code is available at \texttt{github.com/block-wise/block-wise}.

\begin{table*}[t]
\centering
\begin{tabular}{c|cc|cc|c}
Train size & seq & seq with reg & multi-lap seq & multi-lap seq with reg & end-to-end \\ \hline 
50000 & 68.74 $\pm$ 0.45 & \textbf{68.79} $\pm$ 0.56 & 69.48 $\pm$ 0.53 & \textbf{69.95} $\pm$ 0.50 & 75.85 $\pm$ 0.70  \\
25000 & 60.48 $\pm$ 0.15 & \textbf{60.59} $\pm$ 0.14 & 61.33 $\pm$ 0.23 & \textbf{61.71} $\pm$ 0.32 & 65.36 $\pm$ 0.31 \\
12500 & 51.64 $\pm$ 0.33 & \textbf{51.74} $\pm$ 0.26 & 51.30 $\pm$ 0.22 & \textbf{51.89} $\pm$ 0.30 & 52.39 $\pm$ 0.97 \\
5000 & 36.37 $\pm$ 0.33 & \textbf{36.40} $\pm$ 0.40 & 33.68 $\pm$ 0.48 & \textbf{34.61} $\pm$ 0.59 & 36.38 $\pm$ 0.31 \\  \bottomrule
\end{tabular}
\caption{Average highest test accuracy (best block) and 95$\%$ confidence interval of 10-1 ResNet models (256 filters) over 10 runs on CIFAR100 with train sets of different sizes and different methods of training: block-wise sequential (seq), block-wise multi-lap sequential (multi-lap seq), both with and without the transport regularization (reg), and end-to-end.}
\label{tab:seq-cifar100-resnet-10-1}
\end{table*}

\begin{table*}[t]
\centering
\begin{tabular}{cc|cc|c}
seq & seq with reg & multi-lap seq & multi-lap seq with reg & end-to-end \\ \hline 
71.47 $\pm$ 0.60 & \textbf{71.98} $\pm$ 0.58 & 73.35 $\pm$ 0.53 & \textbf{73.59} $\pm$ 0.52 & 75.85 $\pm$ 0.70  \\ \bottomrule
\end{tabular}
\caption{Average highest test accuracy (best block) and 95$\%$ confidence interval of 2-5 ResNet models (256 filters) over 10 runs on CIFAR100 with different methods of training: module-wise sequential (seq), module-wise multi-lap sequential (multi-lap seq), both with and without the transport regularization (reg), and end-to-end.}
\label{tab:seq-cifar100-resnet-2-5}
\end{table*}

The first task is training a 10-block ResNet block-wise on CIFAR100 \cite{cifar} with standard data augmentation. The network starts with an encoder, i.e. a first layer that downsamples the images into a $16 \times 16$ shape with 256 filters and that is also trained greedily with its auxiliary classifier, but without transport regularization. A further downsampling into shape $512 \times 8 \times 8$ takes place after 5 blocks via a convolutional layer that is considered part of the following block but is also not transport regularized. We use orthogonal initialization \cite{orth} with a gain of 0.05. For sequential and multi-lap sequential training, we use SGD with a learning rate of 0.007. For parallel training we use SGD with learning rate of 0.003. For sequential training, block $k$ is trained for $50 + 10 k$ epochs where $0 \leq k \leq 10$, block 0 being the encoder. This idea of increasing the number of epochs per layer along with the depth is found in \cite{cascade}. For multi-lap sequential training, block $k$ is trained for $10 + 2 k$ epochs, and this is repeated for 5 laps. For parallel training, the network is trained for 300 epochs. These architectural and training choices have been made to improve the baseline test performance of vanilla block-wise training without transport regularization. For reference, we also report the end-to-end test performance of the same architecture, trained for 300 epochs with a learning rate of 0.1 that is divided by five at epochs 120, 160 and 200. For each block-wise training method, we report the highest test accuracy achieved along the depth of the greedily trained blocks, for different sizes of the train set. 

In Table \ref{tab:seq-cifar100-resnet-10-1} we report the results for both methods of sequential training and end-to-end training for comparison. We see that multi-lap sequential training improves the test accuracy of sequential training by around $0.8$ percentage points when the training dataset is large, but works less well on a small training set. Our method mainly improves the test accuracy of multi-lap sequential  training. The improvement increases as the training set gets smaller and reaches 1 percentage point. In Table \ref{tab:par-cifar100-resnet-10-1} we report the results for parallel training, which already performs quite close to end-to-end training in the full data regime and even better in the small data regime. Again the improvement in test accuracy from the regularization happens mostly with smaller training sets. As in \cite{belilovsky2}, parallel greedy training performs better than sequential training for the same architecture and a somewhat shorter total training time. An observation that is confirmed in all subsequent experiments. Note that performances of around $70\%$ in these tables for block-wise training on the full CIFAR100 dataset are comparable to, and even a little better than, the DDG method (Table 3 in \cite{ddg}) and the ResIST method (Table 2 in \cite{resist}), even though they only split much deeper ResNets in two parts (DDG) or in up to 8 parts (ResIST).

We also report results for similar experiments with some variations in the architecture and on other datasets (MNIST \cite{mnist} and CIFAR10 \cite{cifar}), but here we report the accuracy achieved by the last block. We see a similar pattern of a greater improvement due to the regularization as the training sets get smaller, gaining as much as 6 percentage points in some cases (Tables \ref{tab:seq-cifar10-resnet-10-1-last}, and \ref{tab:par-mnist-resnet-20-1} and \ref{tab:par-cifar10-resnet-20-1} in the Appendix). We notice that the improvement from the regularization is more important in these three tables where the architecture has not been particularly adapted to block-wise training (i.e. the networks are deeper and not very wide, a classic classifier made up of one or two linear layers is used as opposed to the classifier from \cite{belilovsky1}). Regularizing can then replace too much architectural fine-tuning, although it still helps, even if a little, in all cases. Finally, we include in Table \ref{tab:seq-cifar10-resnet-10-1-best} in the Appendix the accuracy achieved by the best block along the depth for the same experiment as in Table \ref{tab:seq-cifar10-resnet-10-1-last} and we notice a more important improvement in the accuracy of the last block than in the accuracy of the best block when using the regularization. The around $88\%$ accuracy on CIFAR10 of sequential training (Tables \ref{tab:seq-cifar10-resnet-10-1-last} and \ref{tab:seq-cifar10-resnet-10-1-best}) is comparable to results for sequential training in Table 2 of \cite{belilovsky1} (with VGG networks of comparable depth and width). We also observe that with the regularization the difference between the accuracy of the last block and that of the best block is smaller than without the regularization. The regularization then helps to train deeper networks block-wise. 

We confirm this through the following experiment. As the network gets deeper (50 blocks trained for 10 epochs), we expect training it block-wise to become more difficult, and the improvement from the regularization increases when looking at the accuracy of the last block for sequential training (Table \ref{tab:seq-cifar100-resnet-50-1-last} in the Appendix). The Appendix also contains in Tables \ref{tab:par-cifar100-resnext} and \ref{tab:seq-cifar100-resnext} results on ResNeXt-50-32$\times$4d \cite{resnext}, which turns out to be difficult to train block-wise.

\begin{table}
\centering
\begin{tabular}{c|cc|c}
Train & par & par with reg & end-to-end \\ \hline
50000 & 72.59 $\pm$ 0.40 & \textbf{72.63} $\pm$ 0.40 & 75.85 $\pm$ 0.70 \\ 
25000 & 64.84 $\pm$ 0.19 & \textbf{65.01} $\pm$ 0.27 & 65.36 $\pm$ 0.31 \\  
12500 & 55.13 $\pm$ 0.24 & \textbf{55.40} $\pm$ 0.35 & 52.39 $\pm$ 0.97 \\  
5000 & 39.45 $\pm$ 0.23 & \textbf{40.36} $\pm$ 0.23 & 36.38 $\pm$ 0.31 \\  \bottomrule
\end{tabular}
\caption{Average highest test accuracy (best block) and 95$\%$ confidence interval of 10-1 ResNet models (256 filters) over 10 runs on CIFAR100 with train sets of different sizes and block-wise parallel (par) training with and without the transport regularization (reg) and end-to-end training.}
\label{tab:par-cifar100-resnet-10-1}
\end{table}

\begin{table*}[t]
\centering
\begin{tabular}{cccccc}
par (ours) & par with reg (ours) & DGL ResNet152 & PredSim ResNet152 & Sedona ResNet101 & Sedona ResNet152 \\ \hline 
59.56 $\pm$ 0.16 & 60.72 $\pm$ 0.44 & 57.64 & 51.76 & 59.12 & 64.10  \\  \bottomrule
\end{tabular}
\caption{Average test accuracy and $95\%$ confidence intervals of 4-4 ResNet models (256 filters) over 5 runs on TinyImageNet with block-wise parallel training (par) and block-wise parallel training with transport regularization (reg), compared to methods DGL, PredSim and Sedona from \cite{sedona} that also split their networks in 4 module-wise-parallel-trained modules.}
\label{tab:par-tinyimagenet-resnet-4-4}
\end{table*}

\begin{figure*}[t]
\centering
\includegraphics[scale=0.107]{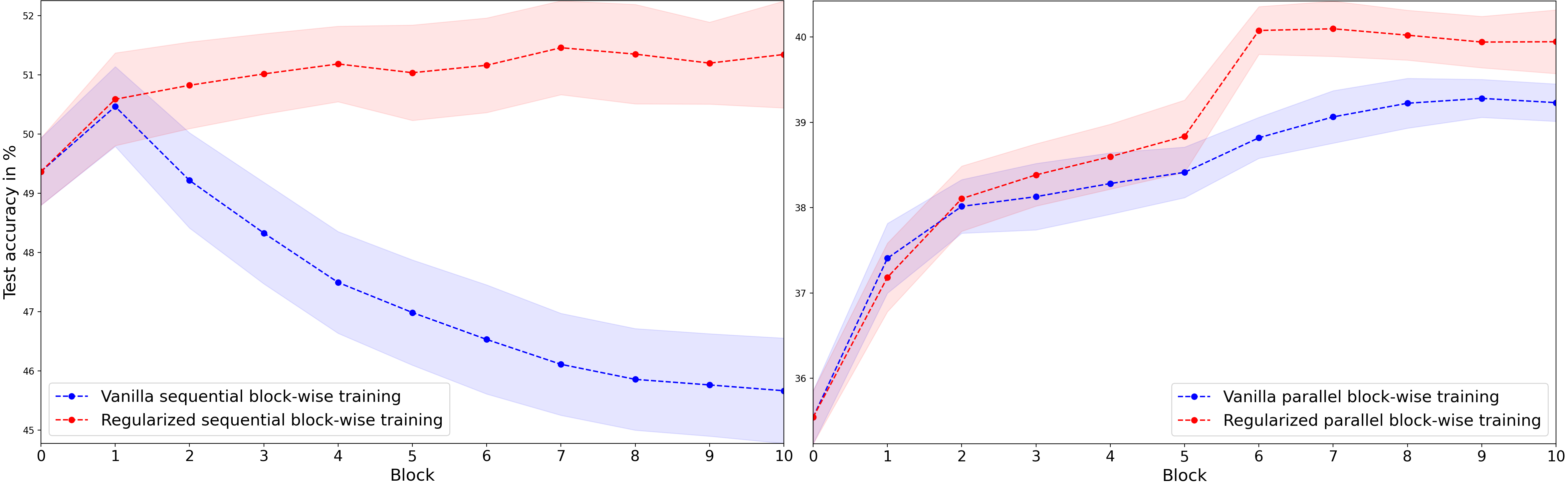}
\caption{Highest test accuracy after each block of 10-block ResNet averaged over 10 runs with $95\%$ confidence intervals. Left: vanilla (blue) and regularized (red) sequential block-wise training on $2\%$ of the CIFAR10 training set. Right: vanilla (blue) and regularized (red) parallel block-wise training on $10\%$ of the CIFAR100 training set.}
\label{fig:seq-par}
\end{figure*}

The second task is splitting the 10-block ResNet from the first task into two modules of 5 blocks (plus the encoder) trained module-wise on all of CIFAR100. When trained sequentially, module $k$ is trained for $50 + 75 k$ epochs for $0 \leq k \leq 2$. When trained sequentially in multiple laps, module $k$ is trained for $10 + 15 k$ epochs for 5 laps. When trained in parallel, we train for 300 epochs. Initialization and learning rates are the same. Here the highest test accuracy along the depth is always that of the second module. In Table \ref{tab:seq-cifar100-resnet-2-5} (sequential training) and Table \ref{tab:par-cifar100-resnet-2-5} in the Appendix (parallel training) we see improvement from the regularization mostly for sequential training. We also find in Table \ref{tab:seq-cifar100-resnet-2-5} that multi-lap sequential training improves by 2 percentage points on simple sequential training. In these tables 2-5 ResNet means 2 modules of 5 blocks each trained module-wise. The $75\%$ test accuracy on CIFAR100 in Table \ref{tab:par-cifar100-resnet-2-5} in the Appendix for parallel training is very close to the end-to-end baseline in Table \ref{tab:seq-cifar100-resnet-10-1}, beats the DDG method (Table 3 in \cite{ddg}) and is comparable to the Features Replay method (Table 2 in \cite{features-replay}). To compare to InfoPro, we train the same model in parallel on CIFAR10 (Table \ref{tab:par-cifar10-resnet-2-5} in the Appendix) and find an improvement of around $0.5$ percentage points over InfoPro (Table 2 in \cite{infopro}).

\begin{table}
\centering
\begin{tabular}{c|cc|c}
Train & seq & seq with reg & end-to-end \\ \hline 
50000 & 88.02 $\pm$ 0.18 & \textbf{88.20} $\pm$ 0.24 & 91.88 $\pm$ 0.18 \\ 
25000 & 83.95 $\pm$ 0.13 & \textbf{84.28} $\pm$ 0.22 & 88.75 $\pm$ 0.27 \\  
10000 & 76.00 $\pm$ 0.39 & \textbf{77.18} $\pm$ 0.34 & 82.61 $\pm$ 0.35 \\  
5000 & 67.74 $\pm$ 0.49 & \textbf{69.67} $\pm$ 0.44 & 73.93 $\pm$ 0.67 \\  
1000 & 45.67 $\pm$ 0.88 & \textbf{51.34} $\pm$ 0.90 & 50.63 $\pm$ 0.98 \\  \bottomrule
\end{tabular}
\caption{Average highest test accuracy (last block) and 95$\%$ confidence interval of 10-1 ResNet models (256 filters) over 10 runs on CIFAR10 with train sets of different sizes and block-wise sequential (seq) training with and without the transport regularization (reg) and end-to-end training.}
\label{tab:seq-cifar10-resnet-10-1-last}
\end{table}

Finally, we train a 16-block ResNet (with 256 initial filters and downsampling and doubling of the filters at the midpoint) divided in 4 modules of 4 blocks (a 4-4 ResNet in our notations) module-wise on TinyImageNet. Parallel module-wise training in this case consumes about $25 \%$ less memory than end-to-end training (6951MiB vs 9241MiB). We compare in Table \ref{tab:par-tinyimagenet-resnet-4-4} our results in this setup to those of three of the best recent parallel module-wise training methods: DGL \cite{belilovsky2}, PredSim \cite{nokland} and Sedona \cite{sedona}, as reported in Table 2 of \cite{sedona}. The benefit of the regularization is clear as it adds 1.2 percentage points of accuracy. While our network has around 51 million parameters, our method easily beats the first two methods, even when they use a bigger ResNet152 (around 59 million parameters) divided in four. Our method beats Sedona when it uses a ResNet101 (43 million parameters) but not a ResNet152. However, Sedona is an architecture search method that first searches for the best auxiliary architectures and the best positions to split the network in four, which requires a long pre-training time, before the actual module-wise training. The methods are therefore not quite comparable and can be combined by adding the regularization at the module-wise training phase of Sedona. We report in Table \ref{tab:seq-tinyimagenet-resnet-4-4} in the Appendix results from training these 4-4 ResNets on TinyImageNet sequentially. We find that multi-lap sequential training improves over simple sequential training by 1 percentage point and that the regularization improves the performance of multi-lap sequential training by almost 1 percentage point.

In Figure \ref{fig:seq-par}, we look at the test accuracy of each block after block-wise training with and without the regularization. On the left, from experiments with sequential block-wise training from Table \ref{tab:seq-cifar10-resnet-10-1-last} on a train set of 1000 CIFAR10 images, we see a large decline in accuracy after the first block (block 0 being the encoder) that our method avoids. On the right, from experiments with parallel block-wise training from Table \ref{tab:par-cifar100-resnet-10-1} on a train set of 5000 CIFAR100 images, we see a steeper increase in test accuracy along the blocks with our method. We see the same pattern in Figure \ref{fig:mls} in the Appendix from experiments with multi-lap sequential block-wise training from Table \ref{tab:seq-cifar100-resnet-10-1} on a train set of 5000 CIFAR100 images.

\section{Conclusion}

We introduced a kinetic energy regularization for block-wise training of ResNets that links block-wise training to gradient flows of the loss in distribution space. The method provably leads to more regular blocks and experimentally improves the test classification accuracy of block-wise sequential, parallel and multi-lap sequential (a variant of sequential training that we introduce) training, especially in small data regimes. The method can easily be adapted to layer-wise training of non-residual networks and combined with other methods that improve the accuracy of layer-wise training. 

Future work can experiment with working in Wasserstein space $W_p$ for $p \neq 2$, i.e. regularizing with a norm $\|.\|_p$ with $p \neq 2$. One can also ask how far the obtained composition network $G_K$ is from being an OT map itself, which could provide a better stability bound than the one obtained by naively chaining the stability bounds \eqref{eq:stability} that follow from each module $T_k$ being an OT map. 

Making layer-wise training competitive allows to benefit from its practical advantages without losing too much performance in comparison with end-to-end training. These benefits include the possibility of on-device training in memory-constrained settings, which can be required for privacy reasons for example, via sequential layer-wise training that only requires loading and training one layer at a time, and training parallelism that differs from and complements data and model parallelism, via parallel layer-wise training.

\bibliography{main}

\appendix

\section{Appendix}

\subsection{Background on optimal transport}\label{appsec:ot}

The Wasserstein space $\mathbb{W}_2(\Omega)$ with $\Omega$ a convex and compact subset of $\mathbb{R}^d$ is the space $\mathcal{P}(\Omega)$ of probability measures over $\Omega$, equipped with the distance $W_2$ given by the solution to the optimal transport problem
\begin{equation}
    \label{eq:kantorovich}
    W_2^2(\alpha, \beta) = \min_{\gamma \in \Pi(\alpha, \beta)} \int_{\Omega \times \Omega} \|x - y \|^2 \diff \gamma(x,y)
\end{equation}
where $\Pi(\alpha, \beta)$ is the set of probability distribution over $\Omega \times \Omega$ with first marginal $\alpha$ and second marginal $\beta$, i.e. $\Pi(\alpha, \beta) = \{ \gamma \in \mathcal{P}(\Omega \times \Omega) \ | \ {\pi_1}_\sharp \gamma = \alpha, \ {\pi_2}_\sharp \gamma = \beta \}$ where $\pi_1(x,y) = x$ and $\pi_2(x,y) = y$. The optimal transport problem can be seen as looking for a transportation plan minimizing the cost of displacing some distribution of mass from one configuration to another. This problem indeed has a solution in our setting and $W_2$ can be shown to be a geodesic distance (see for example \cite{santambrogio2015,villani}). If $\alpha$ is absolutely continuous and $\partial \Omega$ is $\alpha$-negligible then the problem in \eqref{eq:kantorovich} (called the Kantorovich problem) has a unique solution and is equivalent to the Monge problem, i.e.
\begin{equation}
    \label{eq:monge}
    W_2^2(\alpha, \beta) = \min_{T \text{ s.t. } T_\sharp \alpha = \beta} \int_{\Omega} \|T(x) - x \|^2 \diff \alpha (x)
\end{equation}
and this problem has a unique solution $T^\star$ linked to the solution $\gamma^\star$ of \eqref{eq:kantorovich} through $\gamma^\star = (\text{id}, T^\star)_\sharp \alpha$. Another equivalent formulation of the optimal transport problem in this setting is the dynamical formulation \cite{brenier}. Here, instead of directly pushing samples of $\alpha$ to $\beta$ using $T$, we can equivalently displace mass, according to a continuous flow with velocity $v_t: \mathbb{R}^d \to \mathbb{R}^d$. This implies that the density $\alpha_t$ at time $t$ satisfies the \textit{continuity equation} ${\partial_t \alpha_t + \nabla \cdot (\alpha_t v_t) =0}$, assuming that initial and final conditions are given by $\alpha_0=\alpha$ and $\alpha_1 = \beta$ respectively. In this case, the optimal displacement is the one that minimizes the total action caused by $v$ :
\begin{align}
    \label{eq:brenier1}
    W_2^2(\alpha, \beta) = & \min_v \int_0^1 \|v_t\|^2_{L^2(\alpha_t)}\diff t \\
    & \text{s.t. } \partial_t \alpha_t + \nabla \cdot (\alpha_t v_t) = 0, \ \alpha_0 = \alpha, \alpha_1= \beta \nonumber
\end{align}

Instead of describing the density's evolution through the continuity equation, we can describe the paths $\phi^x_t$ taken by particles at position $x$ from $\alpha$ when displaced along the flow $v$. Here $\phi^x_t$ is the position at time $t$ of the particle that was at $x \sim \alpha$ at time 0. The continuity equation is then equivalent to $\partial_t\phi_t^x = v_t(\phi_t^x)$. See chapters 4 and 5 of \cite{santambrogio2015} for details. Rewriting the conditions as necessary, Problem \eqref{eq:brenier1} becomes
\begin{align}
    \label{eq:brenier2}
    W_2^2(\alpha, \beta) = & \min_v \int_0^1 \|v_t\|_{L^2((\phi^\cdot_t)_\sharp \alpha)}^2 \diff t \\
    & \text{s.t. } \partial_t\phi_t^x = v_t(\phi_t^x), \ \phi^\cdot_0 = \text{id}, (\phi^\cdot_1)_\sharp \alpha = \beta \nonumber
\end{align}
and the optimal transport map $T^\star$ that solves \eqref{eq:monge} is in fact $T^\star(x) = \phi_1^x$ for $\phi$ that solves the continuity equation together with the optimal $v^\star$ from \eqref{eq:brenier2}. We refer to \cite{santambrogio2015,villani} for these results on optimal transport.

Optimal transport maps have some regularity properties under some boundedness assumptions. We mention the following result from \cite{figalli}:

\begin{theorem}
\label{th:regularity}
Let $\alpha$ and $\beta$ be absolutely continuous measures on $\mathbb{R}^d$ and $T$ the optimal transport map between $\alpha$ and $\beta$ for the Euclidean cost. Suppose there are bounded open sets $X$ and $Y$, such that the density of $\alpha$ (respectively of $\beta$) is null on $X^\mathsf{c}$ (respectively $Y^\mathsf{c}$) and bounded away from zero and infinity on $X$ (respectively $Y$).

Then there exists two relatively closed sets of null measure $A \subset X$ and $B \subset Y$, such that $T$ is $\eta$-Hölder continuous from $X\setminus A$ to $Y\setminus B$, i.e. $\forall \ x,y \in X \setminus A$ we have $$\|T(x)-T(y)\|\leq C\|x-y\|^\eta \text{ for constants } \eta, C >0$$
\end{theorem}

\subsection{Proof of Theorem 1}\label{appsec:proof}

\begin{proof}
Take a minimizing sequence $(\Tilde{F}_i, \Tilde{T}_i)$, i.e. such that $\mathcal{C}(\Tilde{F}_i, \Tilde{T}_i) \rightarrow \min \mathcal{C}$, where $\mathcal{C} \geq 0$ is the target function in \eqref{eq:nn-wass-min-movement-joint} and denote $\beta_i = \Tilde{T_{i}}_\sharp \rho_k^{\tau}$. Then by compacity $\Tilde{F}_i \rightarrow F^\star$ and $\beta_i \rightharpoonup \beta^\star$ in duality with $\mathcal{C}_b(\Omega)$ by Banach-Alaoglu. There exists $T^\star$ an optimal transport map between $\rho_k^{\tau}$ and $\beta^\star$. Then $\mathcal{C}(F^\star, T^\star) \leq \lim \mathcal{C}(\Tilde{F}_i, \Tilde{T}_i) = \min \mathcal{C}$ by continuity of $\mathcal{L}$ and because 
\begin{align*}
    \int_{\Omega} \| T^\star(x) - x \|^2 \diff \rho_k^{\tau}(x) &= W_2^2(\rho_k^{\tau}, \beta^\star) \\
    &= \lim W_2^2(\rho_k^{\tau}, \beta_i) \\
    &\leq \lim \int_{\Omega} \| \Tilde{T}_i(x) - x \|^2 \diff \rho_k^{\tau}(x)
\end{align*}
as $W_2$ metrizes weak convergence of measures. We take $(F_k^{\tau}, T_k^{\tau}) = (F^\star, T^\star)$. It is also immediate that for any minimizing pair, the transport map has to be optimal. Taking a minimizing sequence $(\Tilde{F}_i, \Tilde{v}^i)$ and the corresponding induced maps $\Tilde{T}_i$ we get the same result for problem \eqref{eq:nn-wass-min-movement-joint-dyn}. The two problems are equivalent by the equivalence between problems \eqref{eq:monge} and \eqref{eq:brenier2}.
\end{proof}

\subsection{Background on gradient flows}\label{appsec:gflow}

We follow \cite{santambrogio2016,ambrosio} for this background on gradient flows. Given a function $\mathcal{L} : \mathbb{R}^d \rightarrow \mathbb{R}$ and an initial point $x_0 \in \mathbb{R}^d$, a \textit{gradient flow} is a curve $x: [0, \infty[ \rightarrow \mathbb{R}^d$ that solves the Cauchy problem
\begin{equation}
\label{eq:cauchy}
\systeme{x'(t) = - \nabla \mathcal{L}(x(t)), x(0) = x_0}
\end{equation}
A solution exists and is unique if $\nabla \mathcal{L}$ is Lipschitz or $\mathcal{L}$ is convex. Given $\tau > 0$ and $x^{\tau}_0 = x_0$ define a sequence $(x^{\tau}_k)_k$ through the \textit{minimizing movement scheme}:
\begin{equation}
\label{eq:euc-min-movement}
    x^{\tau}_{k+1} \in \arg \min_{x \in \mathbb{R}^d}  \ \mathcal{L}(x) + \frac{1}{2 \tau} \| x - x^{\tau}_k \|^2
\end{equation}
$\mathcal{L}$ lower semi-continous and $\mathcal{L}(x) \geq C_1 - C_2 \| x \|^2$ guarantees existence of a solution of \eqref{eq:euc-min-movement} for $\tau$ small enough. $\mathcal{L}$ $\lambda$-convex meets these conditions and also provides uniqueness of the solution because of strict convexity of the target. See \cite{santambrogio2015,santambrogio2016,ambrosio}.

We interpret the point $x^{\tau}_k$ as the value of a curve $x$ at time $k \tau$. We can then construct a curve $x^{\tau}$ as the piecewise constant interpolation of the points $x^{\tau}_k$. We can also construct a curve $\Tilde{x}^{\tau}$ as the affine interpolation of the points $x^{\tau}_k$.

If $\mathcal{L}(x_0) < \infty$ and $\inf \mathcal{L} > - \infty$ then $(x^{\tau})$ and $(\Tilde{x}^{\tau})$ converge uniformly to the same curve $x$ as $\tau$ goes to zero (up to extracting a subsequence). If $\mathcal{L}$ is $\mathcal{C}^1$, then the limit curve $x$ is a solution of \eqref{eq:cauchy} (i.e. a gradient flow of $\mathcal{L}$). If $\mathcal{L}$ is not differentiable then $x$ is solution of the problem defined using the subdifferential of $\mathcal{L}$, i.e. $x$ satisfies $x'(t) \in - \partial \mathcal{L}(x(t))$ for almost every $t$.

If $\mathcal{L}$ is $\lambda$-convex with $\lambda > 0$, then the solution to \eqref{eq:cauchy} converges exponentially to the unique minimizer of $\mathcal{L}$ (which exists by coercivity). So taking $\tau \rightarrow 0$ and $k \rightarrow \infty$, we tend towards the minimizer of $\mathcal{L}$.

The advantage of the minimizing movement scheme \eqref{eq:euc-min-movement} is that it can be adapted to metric spaces by replacing the Euclidean distance by the metric space's distance. In the (geodesic) metric space $\mathbb{W}_2(\Omega)$ with $\Omega$ convex and compact, for $\mathcal{L} : \mathbb{W}_2(\Omega) \rightarrow \mathbb{R} \cup \{ \infty \}$ lower semi-continuous for the weak convergence of measures in duality with $\mathcal{C}(\Omega)$ (equivalent to lower semi-continuous with respect to the distance $W_2$) and $\rho^{\tau}_0 = \rho_0 \in \mathcal{P}(\Omega)$, the minimizing movement scheme \eqref{eq:euc-min-movement} becomes
\begin{equation}
\label{eq:wass-min-movement-app}
    \rho^{\tau}_{k+1} \in \arg \min_{\rho \in \mathcal{P}(\Omega)} \ \mathcal{L}(\rho) + \frac{1}{2 \tau} W_2^2(\rho, \rho^{\tau}_k)
\end{equation}
This problem has a solution because the objective is lower semi-continuous and the minimization is over $\mathcal{P}(\Omega)$ which is compact by Banach-Alaoglu. 

We can construct a piecewise constant interpolation between the measures $\rho^{\tau}_k$, or a geodesic interpolation where we travel along a geodesic between $\rho^{\tau}_k$ and $\rho^{\tau}_{k+1}$ in $\mathbb{W}_2(\Omega)$, constructed using the optimal transport map between these measures. Again, if $\mathcal{L}(x_0) < \infty$ and $\inf \mathcal{L} > - \infty$ then both interpolations converge uniformly to a limit curve $\Tilde{\rho}$ as $\tau$ goes to zero. Under further conditions on $\mathcal{L}$, mainly $\lambda$-geodesic convexity (i.e. $\lambda$-convexity along geodesics) for $\lambda > 0$, we can prove stability and convergence of $\Tilde{\rho}(t)$ to a minimizer of $\mathcal{L}$ as $t \rightarrow \infty$, see \cite{santambrogio2015,santambrogio2016,ambrosio}.

\subsection{Additional experiments}\label{appsec:exp}

\begin{table}[H]
\centering
\begin{tabular}{c|cc|c}
Train & seq & seq with reg & end-to-end \\ \hline  
50000 & 88.14 $\pm$ 0.14 & \textbf{88.34} $\pm$ 0.22 & 91.88 $\pm$ 0.18 \\  
25000 & 84.15 $\pm$ 0.17 & \textbf{84.46} $\pm$ 0.22 & 88.75 $\pm$ 0.27 \\  
10000 & 76.62 $\pm$ 0.40 & \textbf{77.47} $\pm$ 0.35 & 82.61 $\pm$ 0.35 \\
5000 & 69.60 $\pm$ 0.43 & \textbf{70.22} $\pm$ 0.50 & 73.93 $\pm$ 0.67 \\ 
1000 & 51.59 $\pm$ 0.91 & \textbf{52.06} $\pm$ 0.71 & 50.63 $\pm$ 0.98 \\ \bottomrule
\end{tabular}
\caption{Average highest test accuracy (best block) and 95$\%$ confidence interval of 10-1 ResNet (256 filters) over 10 runs on CIFAR10 with train sets of different sizes and block-wise sequential (seq) training with and without the transport regularization (reg) and end-to-end training.}
\label{tab:seq-cifar10-resnet-10-1-best}
\end{table}

\begin{table}[H]
\centering
\begin{tabular}{c|cc|c}
Train & par & par with reg & end-to-end \\ \hline 
60000 & 99.07 $\pm$ 0.04 & \textbf{99.08} $\pm$ 0.04 & 99.30 $\pm$ 0.03 \\ 
30000 & 98.90 $\pm$ 0.05 & \textbf{98.93} $\pm$ 0.06 & 99.22 $\pm$ 0.03 \\ 
12000 & 98.52 $\pm$ 0.06 & \textbf{98.59} $\pm$ 0.06 & 98.96 $\pm$ 0.06 \\ 
6000 & 98.05 $\pm$ 0.09 & \textbf{98.16} $\pm$ 0.07 & 98.62 $\pm$ 0.06 \\
1500 & 96.34 $\pm$ 0.12 & \textbf{96.91} $\pm$ 0.07 & 97.19 $\pm$ 0.08 \\ 
1200 & 95.80 $\pm$ 0.12 & \textbf{96.58} $\pm$ 0.09 & 96.88 $\pm$ 0.09 \\
600 & 91.35 $\pm$ 0.99 & \textbf{95.16} $\pm$ 0.15 & 95.30 $\pm$ 0.17 \\ 
300 & 89.81 $\pm$ 0.73 & \textbf{92.86} $\pm$ 0.24 & 92.87 $\pm$ 0.28 \\ 
150 & 81.84 $\pm$ 1.22 & \textbf{87.48} $\pm$ 0.42 & 87.82 $\pm$ 0.59 \\ \bottomrule
\end{tabular}
\caption{Average highest test accuracy (last block) and 95$\%$ confidence interval of 20-1 ResNet (32 filters, fixed encoder, same classifier) over 20/50 runs on MNIST with block-wise parallel (par) training with and without the transport regularization (reg) and end-to-end training.}
\label{tab:par-mnist-resnet-20-1}
\end{table}

\begin{table}[H]
\centering
\begin{tabular}{c|cc|c}
Train & par & par with reg & end-to-end \\ \hline  
50000 & 85.98 $\pm$ 0.28 & \textbf{86.02} $\pm$ 0.26 & 93.11 $\pm$ 0.19 \\ 
25000 & 80.94 $\pm$ 0.25 & \textbf{81.09} $\pm$ 0.32 & 89.10 $\pm$ 0.29 \\
10000 & 72.49 $\pm$ 0.46 & \textbf{73.01} $\pm$ 0.31 & 80.52 $\pm$ 0.46 \\ 
5000 & 62.31 $\pm$ 0.54 & \textbf{64.06} $\pm$ 0.57 & 69.44 $\pm$ 0.88 \\
500 & 38.61 $\pm$ 0.47 & \textbf{41.44} $\pm$ 0.44 & 40.40 $\pm$ 0.60 \\ \bottomrule
\end{tabular}
\caption{Average highest test accuracy (last block) and 95$\%$ confidence interval of 20-1 ResNet (100 filters, fixed encoder, same classifier) over 10 runs on CIFAR10 with block-wise parallel (par) training with and without the transport regularization (reg) and end-to-end training.}
\label{tab:par-cifar10-resnet-20-1}
\end{table}

\begin{table}[H]
\centering
\begin{tabular}{cc|c}
par & par with reg & end-to-end \\ \hline
75.23 $\pm$ 0.51 & \textbf{75.37} $\pm$ 0.49 & 75.85 $\pm$ 0.70 \\ \bottomrule
\end{tabular}
\caption{Average highest test accuracy and 95$\%$ confidence interval of 2-5 ResNet (256 filters) over 10 runs on CIFAR100 with block-wise parallel (par) training with and without the transport regularization (reg) and end-to-end training.}
\label{tab:par-cifar100-resnet-2-5}
\end{table}

\begin{table}[H]
\centering
\begin{tabular}{cc|c}
par & par with reg & end-to-end \\ \hline
93.90 $\pm$ 0.13  & \textbf{93.93} $\pm$ 0.15 & 94.10 $\pm$ 0.34 \\ \bottomrule
\end{tabular}
\caption{Average highest test accuracy and 95$\%$ confidence interval of 2-5 ResNet (256 filters) over 10 runs on CIFAR10 with block-wise parallel (par) training with and without the transport regularization (reg) and end-to-end training.}
\label{tab:par-cifar10-resnet-2-5}
\end{table}

\begin{table}[H]
\centering
\begin{tabular}{cc|c}
par & par with reg & end-to-end \\ \hline
57.86 $\pm$ 0.49 & \textbf{57.93} $\pm$ 0.51 & 72.97 $\pm$ 1.18 \\ \bottomrule
\end{tabular}
\caption{Average highest test accuracy (best block) and 95$\%$ confidence interval of ResNeXt over 10 runs on CIFAR100 with block-wise parallel (par) training with and without the transport regularization (reg) and end-to-end training.}
\label{tab:par-cifar100-resnext}
\end{table}

\begin{table*}[t]
\centering
\begin{tabular}{cc|cc|c}
seq & seq with reg & multi-lap seq & multi-lap seq with reg & end-to-end \\ \hline 
52.29 $\pm$ 0.53 & \textbf{52.42} $\pm$ 0.65 & 52.59 $\pm$ 0.63 & \textbf{52.84} $\pm$ 0.65 & 72.97 $\pm$ 1.18 \\ \bottomrule
\end{tabular}
\caption{Average highest test accuracy (best block) and 95$\%$ confidence interval of ResNeXt over 10 runs on CIFAR100 with different methods of training: block-wise sequential (seq), block-wise multi-lap sequential (multi-lap seq), both with and without the transport regularization (reg), and end-to-end.}
\label{tab:seq-cifar100-resnext}
\end{table*}

\begin{table*}[t]
\centering
\begin{tabular}{cc|cc|c}
seq & seq with reg & multi-lap seq & multi-lap seq with reg & end-to-end \\ \hline 
63.40 $\pm$ 0.46 & \textbf{63.86} $\pm$ 0.56 & 62.59 $\pm$ 0.64 & \textbf{63.24} $\pm$ 0.50 & 63.34 $\pm$ 2.41 \\ \bottomrule
\end{tabular}
\caption{Average highest test accuracy (last block) and 95$\%$ confidence interval of 50-1 ResNet (256 filters) over 10 runs on CIFAR100 with different methods of training: block-wise sequential (seq), block-wise multi-lap sequential (multi-lap seq), both with and without the transport regularization (reg), and end-to-end.}
\label{tab:seq-cifar100-resnet-50-1-last}
\end{table*}

\begin{table*}[t]
\centering
\begin{tabular}{cc|cc|c}
seq & seq with reg & multi-lap seq & multi-lap seq with reg & end-to-end \\ \hline 
55.53 $\pm$ 1.39 & \textbf{55.58} $\pm$ 1.27 & 56.57 $\pm$ 0.08 & \textbf{57.39} $\pm$ 0.53 & 64.27 $\pm$ 1.28  \\ \bottomrule
\end{tabular}
\caption{Average highest test accuracy and 95$\%$ confidence interval of 4-4 ResNet models (256 filters) over 3 runs on TinyImageNet with different methods of training: block-wise sequential (seq), block-wise multi-lap sequential (multi-lap seq), both with and without the transport regularization (reg), and end-to-end.}
\label{tab:seq-tinyimagenet-resnet-4-4}
\end{table*}

\begin{figure*}[t]
\centering
\includegraphics[scale=0.21]{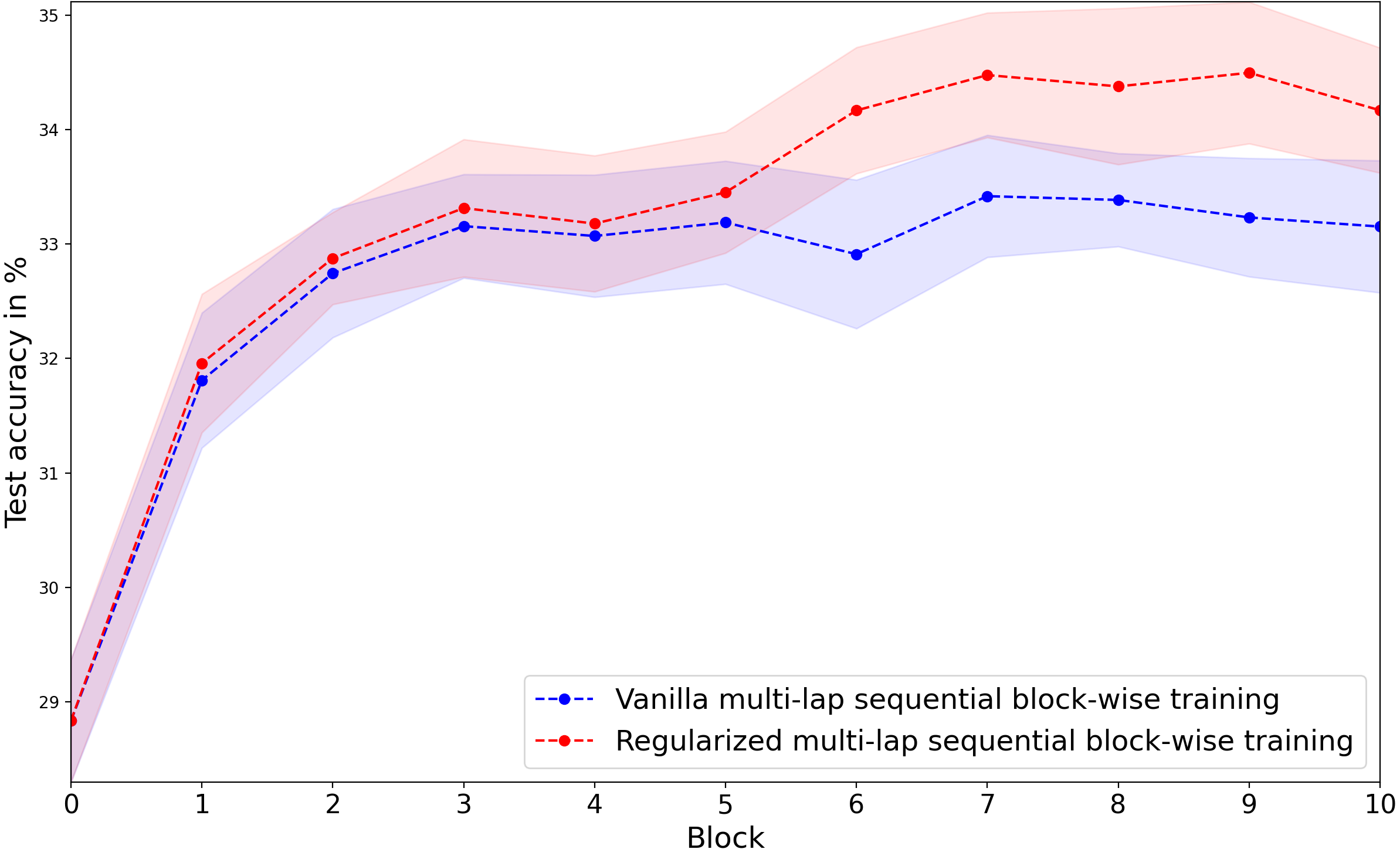}
\caption{Highest test accuracy after each block of 10-block ResNet with multi-lap sequential block-wise training with (red) and without (blue) the transport regularization on $10\%$ of the CIFAR100 training set averaged over 10 runs with $95\%$ confidence intervals.}
\label{fig:mls}
\end{figure*}

\end{document}